%% file: main.tex
\documentclass{ifacconf}
\usepackage{graphicx}      % include this line if your document contains figures
\usepackage{natbib}        % required for bibliography
\usepackage{url}            % simple URL typesetting
\usepackage{booktabs}       % professional-quality tables
\usepackage{amsfonts}       % blackboard math symbols
\usepackage{nicefrac}       % compact symbols for 1/2, etc.
\usepackage{microtype}      % microtypography
\usepackage{xcolor}         % colors
\usepackage{amsmath}
\usepackage{bm}
\usepackage{multicol, multirow}
\usepackage{subcaption}
\usepackage{array}  
\input{macros}

\begin{document}
\begin{frontmatter}

\title{Time-Varying Deep State Space Models for Sequences with Switching Dynamics\thanksref{footnoteinfo}} 

\thanks[footnoteinfo]{
S. Karilanova acknowledges the support of Center for Interdisciplinary Mathematics (CIM), Uppsala University. A.~Özçelikkale acknowledges the 
 support from the Swedish Research Council through grant agreement no. 
2024-05194. 
The computations were enabled by resources provided by the National Academic Infrastructure for Supercomputing in Sweden (NAISS), partially funded by the Swedish Research Council through grant agreement no. 2022-06725. 
}

\author{Sanja~Karilanova, Subhrakanti~Dey, Ayça~Özçelikkale}
\address{Department of Electrical Engineering, Uppsala University, Sweden, (e-mail: \{Sanja.Karilanova, Subhrakanti.Dey, Ayca.Ozcelikkale\}@angstrom.uu.se).}

\begin{abstract}                % Abstract of not more than 250 words.
The identification and modeling of time-varying systems is a fundamental challenge in signal processing and system identification. To address this challenge, we propose a class of time-varying state-space model (SSM) based neural networks in which the neurons' states are governed by time-varying dynamics. The proposed model provides the learnable time-varying dynamics through a dictionary of basis functions, where each basis function evolves differently over time. 
We evaluate the proposed approach on both synthetic data from switching systems and a speech denoising task where real audio is corrupted with switching dynamics noise. The results show that the proposed time-varying model consistently outperforms its time-invariant counterparts while maintaining comparable computational complexity. 
Our investigations also reveal which aspects of the time-varying dynamics of the data most need to be captured by the proposed time-invariant models, how the additional freedom provided by time-varying basis functions should be allocated across model components, and to what extent larger models can compensate for time-invariant limitations.
\end{abstract}


\end{frontmatter}
%===============================================================================

\section{Introduction}
% Visualization of time-varying matrix parameters
\begin{figure*}
  \centering
  \includegraphics[width=0.8 \linewidth]{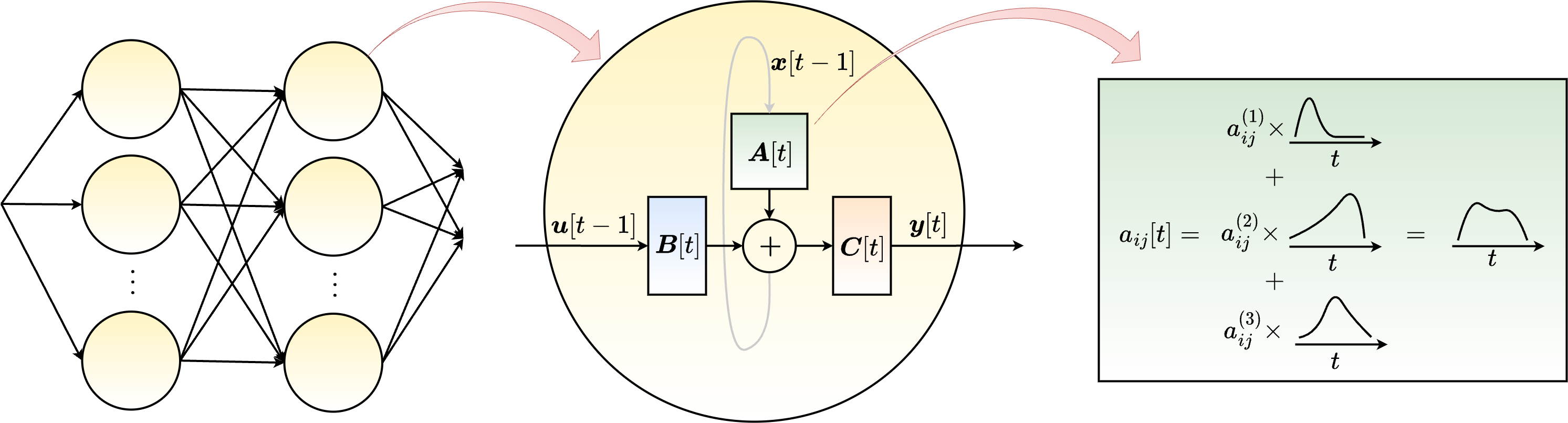} 
  \caption{Left:  Example network architecture with an input layer with a single channel, two hidden layers with SSM neurons and an output layer of two channels. Middle:  An SSM neuron with time-varying dynamics. Right:  Illustration  of the  proposed basis function expansion for a single time-varying element of the SSM state transition matrix.}
  \label{fig:main}
\end{figure*}

% Time varying data 
Many real-world processes are inherently time-varying, meaning that their underlying dynamics evolve over time. This time-varying behavior appears in various domains, such as financial markets, weather forecasting, energy consumption, and neural activity analysis. A central challenge in system identification is therefore to build models that are flexible enough to capture such evolving dynamics \cite{10164274, liu2023nonstationarytransformersexploringstationarity}.

% SSMs
State Space Models (SSMs) provide a principled and interpretable framework for system identification. In their classical form with linear state transitions, often referred to as Linear Dynamical Systems (LDS) \cite{Kalman}, they assume fixed time-invariant transition dynamics. 

% Switched SSMs / SLDS 
A common extension of LDSs for handling time-varying data is the Switching Linear Dynamical System (SLDS) \cite{sun2005switched}, where the dynamics transition between multiple LDS modes.
% Limitations of SLDS
While effective in capturing regime changes, learning SLDSs remains challenging, as it requires estimating the number of modes, the model order of each mode, and the corresponding parameters \cite{PAOLETTI2007242}. 

% Deep-SSM
Another extension of LDS, distinct from SLDS, is deep-SSMs. These models stack LDS blocks into layers with nonlinear transformations and have recently achieved performance comparable to transformers on long-sequence modeling tasks \cite{gu2022efficientlymodelinglongsequences} \cite{smith2023simplified}, as well as have been applied to nonlinear system identification \cite{GEDON2021481}. Extensions to Deep-SSMs include input-dependent mechanism \cite{gu2024mamba} and input timescale invariant formulation \cite{gu2020hippo}.

% SP and Control
Traditionally, system identification research has long explored time-varying and non-stationary behavior through methods such as adaptive filtering, basis function expansions \cite{258089, 1164152, 7254, Zou2003}, and change point detection \cite{Burg2020AnEO}. 
% Time-Varying Neural Networks
More recently, deep learning has also become a powerful tool for system identification \cite{LJUNG20201175}, including approaches based on time-varying neural networks, such as dynamic neural networks \cite{10164274} and non-stationary transformer architectures \cite{liu2023nonstationarytransformersexploringstationarity}. 

% Our work motivation
Our work brings the time-varying basis function expansion methodology into contemporary deep SSM architectures and illustrates the performance for system identification tasks. 
Unlike explicit modeling through SLDSs, our framework requires no explicit specification of switching modes and instead models smooth, continuously time-varying dynamics, enabling efficient training through standard backpropagation through time (BPTT).
In contrast to modeling time variation in deep SSMs through input-dependence, our framework explicitly models time-dependent dynamics independent of the input. Figure \ref{fig:main} provides a visualization of our proposed framework.

% Contributions
The key contributions of this paper are as follows:
\begin{itemize}
    \item We propose a novel time-varying deep SSM framework using learnable basis functions to parameterize state transition, input, and output matrices over time.
    \item We provide experimental validation on both synthetic and real-world data on audio denoising scenarios with non-stationary noise dynamics.
    \item We explore the trade-offs in the proposed model architecture, including effects of different parts of SSM dynamics and basis function allocation,  and compare it with its time-invariant counterpart.
\end{itemize}
Overall, our results show that the proposed time-varying model 
consistently outperforms its time-invariant counterparts under switching dynamics data while retaining comparable computational complexity.

\section{Methods}

\subsection{Preliminaries}

\subsubsection{Time-invariant SSM model:}
\label{sec:time-invariant-ssm}
A time-invariant linear discrete-time SSM is given by \cite{ljung_SI}
\begin{subequations}
\label{eqn:SSM:TInv}
\begin{align}
\label{eqn:SSM:TInv:state}
    \xbf[t] &= \Abf \xbf[t-1] + \Bbf \ubf[t-1] \\
\label{eqn:SSM:TInv:output}
    \ybf[t] &= \Cbf \xbf[t], 
\end{align}
\end{subequations}
with possibly learnable matrix parameters
$\Abf \in \R^{\Nstate \times \Nstate}$, $\Bbf \in \R^{\Nstate \times \nin}, \Cbf \in \R^{\nout\times \Nstate}, \Dbf \in \R^{\nout\times \nin}$, input $\ubf[t] \in \R^{\nin \times 1}$, state variable $\xbf[t] \in \R^{\Nstate \times 1}$, and output $\ybf[t] \in \R^{\nout \times 1}$. 

We refer to the SSM in \eqref{eqn:SSM:TInv} as a neuron in the context of a SSM based neural network. 
Multiple neurons with such SSM dynamics may be used to form a SSM layer, see Figure~\ref{fig:main}. Layers with SSM
dynamics are combined with possibly non-linear activation function layers and mixing layers to form a deep SSM network \cite{gu2022efficientlymodelinglongsequences} \cite{smith2023simplified}.

\subsubsection{Basis function expansion:}
Basis function expansion is a way of representing a function, $f(t)$, as a linear combination of simpler functions, $\fo^{(k)}(\cdot)$, called basis functions, as follows:
\begin{align}
    f(t) = \sum^{\Nbasefnc}_{k=1} \alpha^{(k)} \times \fo^{(k)}(t),
\end{align}
where $\alpha_k \in \R$ are the coefficients. 
Basis functions may assume various forms, such as with Fourier series, Gaussian functions, polynomial functions, rational orthogonal functions, and radial basis functions.
For example, we can have
% $\fo^{(k)}=1$ constant function, 
$\fo^{(k)}=\Norm(t|\mu_k,\sigma_k^2)$  for a dictionary of Gaussian functions with $\Norm(\cdot)$ representing the shape of a Gaussian function and 
$\fo^{(k)}=\sin(w_k t+ \phi_k)$ for a dictionary of sine functions,  
where $\mu_k, \sigma_k, w_k, \phi_k$ are fixed pre-defined scalars. 

\subsection{Proposed time-varying SSM model}
\label{sec:time-vary-model}

We propose the following time-varying SSM model
\begin{subequations}
\label{eqn:SSM:TVar}
\begin{align}
\label{eqn:SSM:TVar:state}
    \xbf[t] &= \Abf[t] \xbf[t-1] + \Bbf[t] \ubf[t-1] \\
\label{eqn:SSM:TVar:output}
    \ybf[t] &= \Cbf[t] \xbf[t],
\end{align}
\end{subequations}
where $\Abf[t], \Bbf[t], \Cbf[t]$ have the same dimensions as the time-invariant model in \eqref{eqn:SSM:TInv}. However, each element of these matrices is now a linear combination of basis functions. In particular, for $\Abf[t]$, we have 
\begin{align}
\label{eqn:aij:basisexpansion}
    [\Abf[t] ]_{ij} = a_{i,j}[t] = \sum^{\Nbasefnc_A}_{k=1} a_{i,j}^{(k)} \times \fo_{A,i,j}^{(k)}[t]
\end{align}
where $[\Abf[t] ]_{ij} =a_{i,j}[t] $ denotes the $i$th row $j$th column element of the matrix $\Abf[t]$. 
An illustration of the expansion in \eqref{eqn:aij:basisexpansion} is provided in Figure~\ref{fig:main}.
We note that the number of basis functions $\Nbasefnc_A$ is independent of any other dimensions in the model. Here we present the general case where the basis function $\fo_{A,i,j}^{(k)}$ may differ across SSM matrices $(A, B, C)$, indices $(i,j)$, and expansion terms $(k)$. In practice, this heterogeneity can be reduced by using a smaller shared dictionary of basis functions.

Similar to $\Abf[t]$, we have the following:
\begin{align}
    [\Bbf[t] ]_{ij} &= b_{i,j}[t] = \sum^{\Nbasefnc_B}_{k=1} b_{i,j}^{(k)} \times \fo_{B,i,j}^{(k)}[t] \\
    [\Cbf[t] ]_{ij} &= c_{i,j}[t] = \sum^{\Nbasefnc_C}_{k=1} c_{i,j}^{(k)} \times \fo_{C,i,j}^{(k)}[t].
\end{align}
 We note that $\Nbasefnc_A $, $\Nbasefnc_B$, 
 $\Nbasefnc_C$ can be chosen independently of each other.
Hence, a subset of $\Abf[t],\Bbf[t],\Cbf[t]$  can remain time-invariant as in the time-invariant model \eqref{eqn:SSM:TInv} independently of the others.

\subsection{Stability}

A central question in deep SSM development is to ensure stability of the SSM dynamics. The stability  of the dynamics in \eqref{eqn:SSM:TVar} is controlled by $\Abf[t]$. 
For simplicity of presentation, we now assume that $\Abf[t]$ is a diagonal matrix and the basis functions satisfy  $|\fo_{A,i,j}^{(k)}[t]|\leq 1$. To ensure stability, the modulus of the diagonal entries (eigenvalues) of $\Abf[t]$ must be strictly smaller than $1$, i.e., $|a_{i,i}[t]|<1$. Using the triangle inequality,  we have 
\begin{subequations}
\begin{align}
    |a_{i,i}[t]| 
    &= |\sum^{\Nbasefnc_A}_{k=1} a_{i,i}^{(k)} \times \fo_{A,i,i}^{(k)}[t]| \\
    & \leq \sum^{\Nbasefnc_A}_{k=1} | a_{i,i}^{(k)} \times \fo_{A,i,i}^{(k)}[t] |
    \leq \sum^{\Nbasefnc_A}_{k=1} |a_{i,i}^{(k)}|
\end{align}
\end{subequations}
Hence, if $\sum^{\Nbasefnc_A}_{k=1} | a_{i,i}^{(k)} | <1$ is guaranteed, then $|a_{i,i}[t]| < 1$, which imply stability of the dynamics governed by $\Abf[t]$.
During training, we enforce the stability condition, $\sum^{\Nbasefnc}_{k=1} | a_{i,i}^{(k)} | <1$, by checking it once per forward pass rather than at each time step. If this condition is violated, i.e., $\sum^{\Nbasefnc}_{k=1} | a_{i,i}^{(k)} | = c  > 1$, then we apply a scaling strategy to enforce the constraint by redefining the coefficients as $\widehat{a}_{i,i}^{(k)}= \frac{1}{c+\epsilon} a_{i,i}^{(k)}$. With this scaling, we have the following that ensure stability 
\begin{align}
    \sum^{\Nbasefnc}_{k=1} | \widehat{a}_{i,i}^{(k)} | 
    = \frac{1}{c+\epsilon} \sum^{\Nbasefnc}_{k=1} | a_{i,i}^{(k)} | 
    = \frac{1}{c+\epsilon} c 
    < 1.
\end{align}

\subsection{Parameter count}
\label{sec:param_count_complexity}

% Params introduced by A,B,C
In \eqref{eqn:SSM:TInv}, each element of $\Abf$, $\Bbf$, $\Cbf$ is a single scalar learnable parameter, while each element of $\Abf[t]$, $\Bbf[t]$ and $\Cbf[t]$ matrices in \eqref{eqn:SSM:TVar} is a linear combination of $\Nbasefnc_A$, $\Nbasefnc_B$ and $\Nbasefnc_C$ basis functions where each coefficient is trainable, hence it corresponds to $\Nbasefnc_A$, $\Nbasefnc_B$ and $\Nbasefnc_C$ learnable parameters per element, respectively. 
Hence, the proposed model increases the trainable parameters associated with $\Abf$, $\Bbf$ and $\Cbf$ by $\Nbasefnc_A$, $\Nbasefnc_B$ and $\Nbasefnc_C$ per neuron, respectively. 

\begin{table}[]
    \centering
    \caption{Trainable parameters in the baseline (time-invariant SSM) and scaling factor relative to the proposed time-varying SSM.}
    \label{tab:nb_params}
    \newcolumntype{l}{>{\centering\arraybackslash}p{2.2cm}}
    \newcolumntype{m}{>{\centering\arraybackslash}p{3.1cm}}
    \newcolumntype{n}{>{\centering\arraybackslash}p{1.7cm}}
    \begin{tabular}{lmn}
    \hline
        Parameter & Nb of trainable params. (Time-Inv.)  & Scaling factor \\
    \hline
        $\Abf$ (Diag) & $\Nstate$ & $\Nbasefnc_A$ \\
        $\Abf$ (non-Diag) & $\Nstate \times \Nstate$ & $\Nbasefnc_A$ \\
        $\Bbf$ & $\nin \times \Nstate$ & $\Nbasefnc_B$ \\
        $\Cbf$ & $\Nstate \times \nout$ & $\Nbasefnc_C$\\
        $\CbfBias$ & $\nout$ & 1\\
        $\Wbf$  & $\Hneurons \times \nout \times \Hneurons \times \nin$ & 1 \\
        Norm. layer & $2 \times \Hneurons \times \nin$ & 1 \\
    \hline
    \end{tabular}
\end{table}

Table~\ref{tab:nb_params} summarizes the parameter counts and scaling factors, where $h$ denotes the number of neurons in a layer governed by either \eqref{eqn:SSM:TInv} or \eqref{eqn:SSM:TVar}; $\Wbf$ denotes the weight matrix for the mixing layer between a SSM layer with $h$ neurons each with $\nout$ outputs and $\nin$ inputs; $\CbfBias$ is a learnable output bias added to \eqref{eqn:SSM:TInv:output} and \eqref{eqn:SSM:TVar:output}.

In many cases, the neural network has  a large number of SSM neurons with a low dimensional state space, i.e., $n \ll \Hneurons$, see,  e.g. neuromorphic computing benchmark \cite{yik2024neurobench}.  Hence,  under moderate dictionary sizes, the additional number of learnable parameters brought by the proposed time-varying model is expected to be low compared to the total model size.   

For $\nout=\nin=1$ and diagonal $\Abf$, the total number of learnable parameters per neuron in the time-invariant case is 
$p_{invary} =  3\Nstate_{invar}+1$, while in the time-varying case is $p_{vary}=\Nstate_{vary}(\Nbasefnc_A+\Nbasefnc_B+\Nbasefnc_C) +1$, where $\Nstate_{invar}$ and $\Nstate_{vary}$ are the state dimension of the neuron in the time-invariant and time-varying case, respectively.
As shown in Section~\ref{sec:results:distorted_speech}, the time-varying model can outperform the time-invariant one even when $p_{vary} = p_{invary}$.

\subsection{Inference complexity}

The learnable coefficients $a_{i,j}^{(k)}, b_{i,j}^{(k)}, c_{i,j}^{(k)}$ are determined during the training based on training data. For the inference phase, i.e. when the model performs a forward pass to evaluate on test data,
the matrices $\Abf(t),\Bbf(t),\Cbf(t)$ can be pre-computed using these coefficients. Hence, during inference, the number of Multiply–Accumulate (MAC) operations is the same for the time-invariant model in \eqref{eqn:SSM:TInv} and the proposed time-varying model \eqref{eqn:SSM:TVar} if the network architecture is otherwise the same.

Hence, even if the number of trainable parameters for the time-varying model is higher than the invariant model for a fixed $n,\nin,\nout$, the inference MAC complexity of the time-varying model is the same. Nevertheless, under this strategy of  pre-computation of $\Abf(t),\Bbf(t),\Cbf(t)$,  the space complexity of the time-invariant model is higher, scaling with sequence length $T$, due to storage of time-dependent matrix parameters.

\section{Experimental set-up}

\subsubsection{Model initialization and training}

% Network architecture
For all experiments in Section~\ref{sec:results}, we use a network with an input layer, one or two hidden layers, and an output layer. The input and output layers correspond to the data channels and the task-specific output channels, respectively. Each hidden layer contains $\Hneurons$ SSM neurons with either identity or The Gaussian Error Linear Unit (GELU) activation. 
% Trainable parameters
The trainable parameters in the networks are the following:
$\Acoeff, \Bcoeff, \Ccoeff$ for each SSM neuron, 
the weight connections between layers $\Wbf$ and the normalization layers' mean and variance scale. 
% Initializations
We initialize $\Bcoeff$ as ones, $\Ccoeff \sim U[0,1)$, $\CbfBias$ as zeros, and $\Acoeff$ using the real part of S4D-Lin initialization~\cite{gu2022parameterization}, expanded across the $\Nbasefnc_A$ dimension by setting $a_{i,j}^{k}=a_{i,j}/\Nbasefnc_A$. We initialize $\Wbf$ uniformly, use batch normalization between layers, and train with BPTT and AdamW. All models are trained on an NVIDIA Tesla T4 GPU (16GB RAM). Further details are in Appendix~\ref{sec:appendix_info_training}.

\subsubsection{Choice of basis functions}
For the basis function expansion for the elements of $\Abf[t], \Bbf[t], \Cbf[t]$, we use dictionary of Gaussian functions and an additional constant function. Specifically, if $\Nbasefnc$ is the number of basis functions and $T$ is the sequence length, then per element we have $1$ constant function and $\Nbasefnc-1$ Gaussian-shaped functions with amplitude $1$, mean $\in U(0, T)$, and standard deviation $\in U(\frac{T}{5(\Nbasefnc-1)+1}, \frac{T}{(\Nbasefnc-1)/3+1})$.

\subsubsection{Performance metrics for evaluation} 
The Mean-Square-Error (MSE) $\in [0,\infty)$ is calculated between the vectors $\sbf$ and $\shatbf$ per time step, where MSE~$ =0$ implies a perfect match i.e. $\sbf = \shatbf$.
Signal-to-noise ratio (SNR) and Scale-invariant signal-to-noise ratio (SI-SNR), are used as defined in \cite{8683855}. Higher dB indicates cleaner speech. While task-dependent,  SI-SNR values $<0$ dB imply the estimate is worse than the noise, $0-5$dB imply low interpretability of speech, $5-15$dB imply decent enhancement and $15-20$dB imply very good enhancement \cite{subakan2022real, wichern2019wham}. All performance values reported use the testing subset of the datasets and are averages over multiple model initializations.

\subsubsection{Trainable parameter count fairness}
In a set of experiments in Section ~\ref{sec:results:distorted_speech}, we match the number of trainable model parameters between time-varying and time-invariant models. Following Section~\ref{sec:param_count_complexity}, we keep the layer architecture the same and for each SSM neuron, we  set $p_{invary}=p_{vary}$, i.e.,  $\Nstate_{invar} = \frac{\Nstate_{vary}}{3}(\Nbasefnc_A+ \Nbasefnc_B + \Nbasefnc_C)$.

\subsubsection{Dataset 1: Four mode system}
\label{sec:illustrative-example-setup}

We generate a dataset from an SLDS with $4$ operating modes, with a fixed switching order and equal mode durations. 
Inputs to the SLDS are linear combination of two sinusoidal sequences with time steps $N=128$, phase $\phi \in U[0, 2\pi]$ and frequency $\omega=\frac{2\pi l}{N}$, where $l\in U[0, N/2]$. The input then passes through the four mode SLDS and creates the output $y[t]$.  
% We make varations 
We consider variations where each of ($\Abf$, $\Bbf$, $\Cbf$) independently switches or remains fixed. 
The dataset consists of $2000$ input-output pairs ($80\%$ train, $20\%$ test), and the task is to reproduce the SLDS output given the input.
Further details are provided in Appendix~\ref{sec:SLDS_dataset}.

\subsubsection{Dataset 2: Distorted speech}
\label{sec:dataset:distorted_speech}
% Visualization of the speech distortion setup
\begin{figure}
  \centering
  \includegraphics[width=\linewidth]{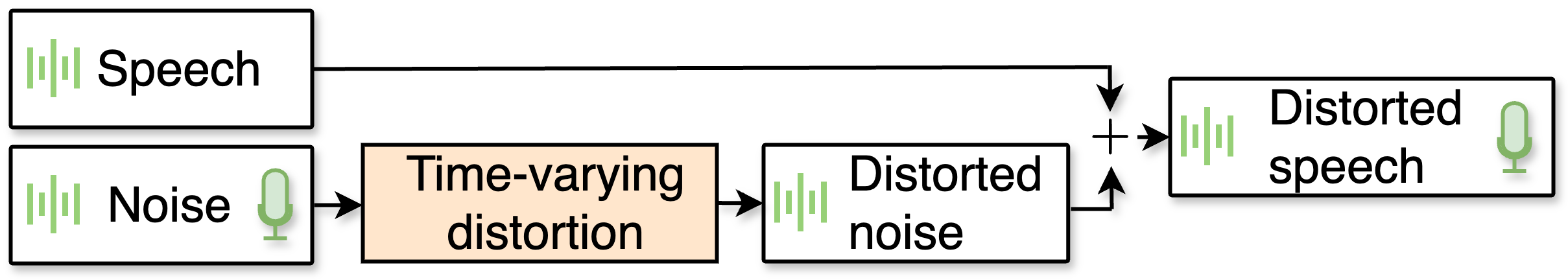} 
  \caption{Visualization of the speech distortion set-up.}
  \label{fig:speech-distortion}
  %\kern-0.4em
\end{figure}

As a real-world case study, we consider audio denoising with time-varying, switched noise dynamics. 
% MSWC-clean
We use the ``surrounding'' keyword subset of the MSWC dataset~\cite{mswcdataset} consisting of $500$ train, $100$ val, and $100$ test samples. Each sample is $1$s long, which is represented with $48000$ discrete-time points. These samples represent clean speech. 
% Noise
We corrupt the clean speech with the four-mode SLDS noise from Section~\ref{sec:illustrative-example-setup}. Figure~\ref{fig:speech-distortion} provides a visualization of the dataset creation.
% Details
Each noise cycle spans $128$ points, giving $375$ repetitions over the $48000$-point speech signal, mimicking recurring noise patterns such as rotating machinery. The initial SNR is set to $5$dB. See Appendix~\ref{sec:setup:DistortedSpeech} for more details.

% Explain task/goal
The model takes the noise as input and learns to estimate the distorted speech; clean speech is then recovered by subtraction. This mirrors active noise cancellation scenarios where the noise signal is accessible but clean speech is not.

\section{Results}
\label{sec:results}

\begin{table*}
  \centering
  % \newcolumntype{n}{>{\centering\arraybackslash}p{1.3cm}}
  % \newcolumntype{m}{>{\centering\arraybackslash}p{0.7cm}}
  % \newcolumntype{l}{>{\centering\arraybackslash}p{0.1cm}}
  \caption{MSE for different Data (columns) and Model (rows) configurations. For the model parameters ($\AbfModel, \BbfModel, \CbfModel$) and data parameters ($\AbfData, \BbfData, \CbfData$), `o' represents time-varying/switched, while `x' represents time-invariant/fixed, respectively. The reported values are rounded averages to one decimal point. Lower values of MSE is better.}
    \label{tab:Data_vs_Model_configurations:mse}
    \begin{tabular}{c|c|cccccccc}
    \hline
    && \multicolumn{8}{c}{Model} \\
    \hline
    &$\Abf\Bbf\Cbf$ & xxx & xxo & xox & xoo & oxx & oxo & oox & ooo \\
    \hline
    \multirow{5}{*}{\rotatebox{90}{Data}} &
    xxx & $0.0$ &$0.0$ &$0.0$ &$0.0$ &$0.0$ &$0.0$ &$0.0$ &$0.0$ \\
    &xxo & $\!\!\!1.7\times 10^{0}$ &$ 7.8\times 10^{-3}$&$ 1.3\times 10^{-2}$&$ 7.1\times 10^{-3}$&$ 4.6\times 10^{-1}$&$ 6.9\times 10^{-3}$&$ 8.1\times 10^{-3}$&$ 6.4\times 10^{-3}$\\
    &xox & $ \!\!\!1.9\times 10^{0}$ &$ 5.0\times 10^{-2}$&$ 2.6\times 10^{-2}$&$ 2.4\times 10^{-2}$&$ 5.1\times 10^{-1}$&$ 1.7\times 10^{-2}$&$ 1.6\times 10^{-2}$&$ 1.3\times 10^{-2}$\\
    &oxx & $4.0\times 10^{-1}$ & $ 1.1\times 10^{-2}$&$ 1.3\times 10^{-2}$&$ 7.6\times 10^{-3}$&$ 4.9\times 10^{-3}$&$ 4.0\times 10^{-3}$&$ 4.0\times 10^{-3}$&$ 3.4\times 10^{-3}$\\
    &ooo &$ 6.1\times 10^{-1 }$&$ 8.0\times 10^{-3 }$&$ 9.0\times 10^{-3 }$&$ 9.0\times 10^{-3 }$&$ 3.9\times 10^{-1 }$&$ 8.0\times 10^{-3 }$&$ 6.0\times 10^{-3 }$&$ 5.0\times 10^{-3 }$\\
    \hline
    \end{tabular}
\end{table*}

\subsection{Four mode system}
\label{sec:results:four_mode_system}
%---------------------------------------------------------------------
\begin{figure}
    \centering
    \includegraphics[width=\columnwidth]{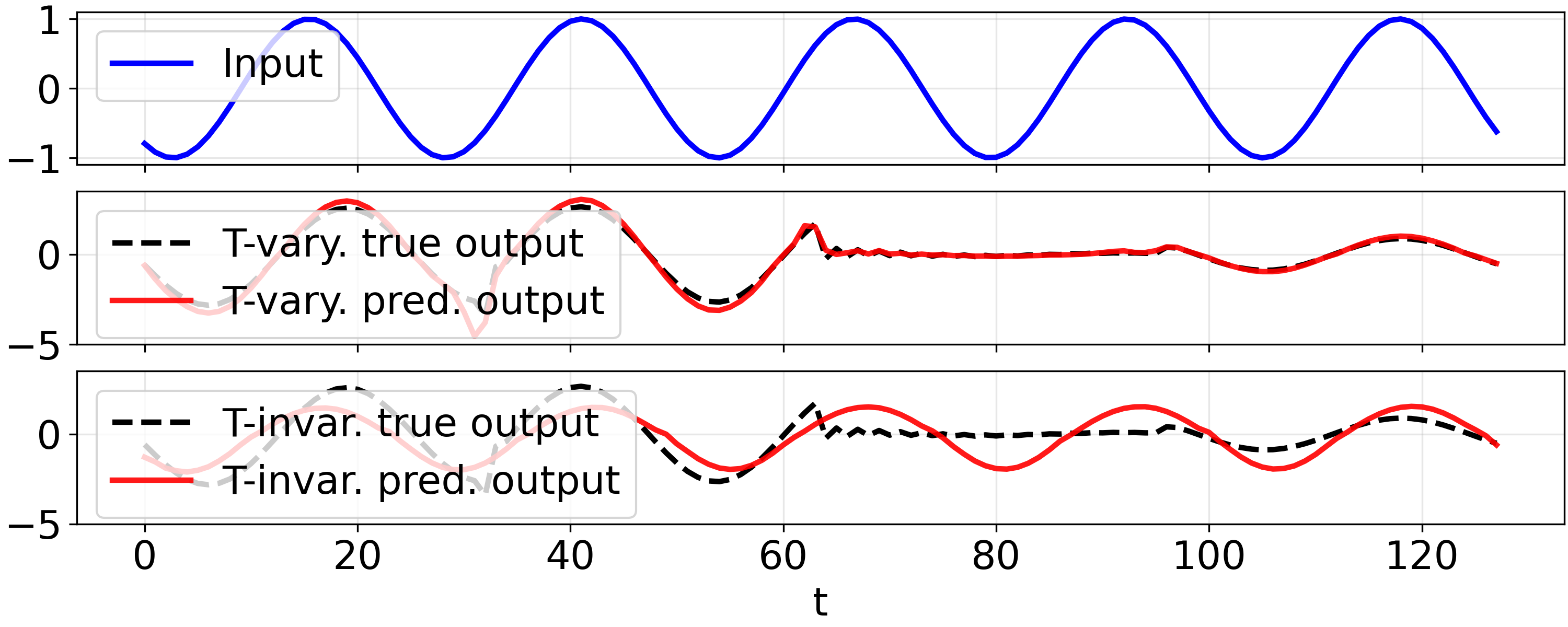}
    \caption{Example sample from the four mode system task.}
    \label{fig:4modesystem_sample_example}
\end{figure}
%---------------------------------------------------------------------
We refer to the data-generating SSM matrices as $\AbfData, \BbfData, \CbfData$ and the network models SSM matrices as $\AbfModel, \BbfModel, \CbfModel$.

% Visualization of samples
Figure~\ref{fig:4modesystem_sample_example} shows a sine input passed through an SLDS with switching $\AbfData, \BbfData, \CbfData$. The time-varying model captures mode changes well, while the time-invariant model learns an average across modes.

% Explain/introduce the combination table
Table \ref{tab:Data_vs_Model_configurations:mse} examines combinations of switched and fixed $\AbfData, \BbfData, \CbfData$ and time-varying and time-invariant model parameters $\AbfModel, \BbfModel, \CbfModel$.
% Observation: 
When none of $\AbfData, \BbfData, \CbfData$ is switched (first column), all models achieve MSE~$=0$. This is consistent with the fact that there is no time-varying dynamics in the data.
% Observation: 
The model with all $\AbfModel, \BbfModel, \CbfModel$  time-varying (last row) matches or outperforms all models across every data configuration. This is consistent with the fact that this is the most flexible model.
% Observation: 
When $\AbfModel$ is time-varying alongside at least one of $\BbfModel$ or $\CbfModel$ (oxo or oox), performance is similar to the fully time-varying case, suggesting both $\BbfModel$ and $\CbfModel$ need not be time-varying simultaneously.
% Observation: 
When only $\AbfModel$ is time-varying (oxx), this model-configuration performs worse than when only $\BbfModel$ or $\CbfModel$ is time-varying (xox or xxo). This suggests that time-varying $\AbfModel$ cannot compensate for switched $\BbfData$ and/or $\CbfData$ since the row `oxx' has higher MSE  values (at the MSE level $10^{-1}$) for all except for its own `oxx' column. On the other hand, the rows `xox' for time-varying $\BbfModel$ and `xxo' for time-varying $\CbfModel$, can obtain relatively low values (at the MSE level $10^{-2}$) for all columns. 

%\kern-0.5em
\subsection{Distorted speech}
\label{sec:results:distorted_speech}
We now discuss the distorted speech results, where `time-varying' and `time-invariant' refer to all of $\AbfModel, \BbfModel, \CbfModel$ being time-varying or time-invariant, respectively.

\subsubsection{Layer depth and activation function:}
\label{sec:distoreted_speech:multi_layer_and_non_linear}

\begin{table*}[]
    \centering
    \caption{SI-SNR for speech denoising task for varying $\Hneurons$ and fixed $\Nstate_{invar}=16,\Nstate_{vary}=4, \Nbasefnc_A=\Nbasefnc_B=\Nbasefnc_C=4$.}
    \label{tab:vary-Hneurons}
    \begin{tabular}{ccccccccc}
        \hline
        \multirow{2}{*}{$\Hneurons$}  
        & \multicolumn{2}{c}{L=1, Identity Act.}
        & \multicolumn{2}{c}{L=2, Identity Act.}
        & \multicolumn{2}{c}{L=1, GELU Act.}
        & \multicolumn{2}{c}{L=2, GELU Act.} \\
        \cline{2-3} \cline{4-5} \cline{6-7} \cline{8-9}
        & Time-Vary. & Time-Invar.
        & Time-Vary. & Time-Invar.
        & Time-Vary. & Time-Invar.
        & Time-Vary. & Time-Invar.\\
        \hline
        64 & $16.1\pm0.5 dB$ & $7.8\pm0.0 dB$
            & $15.8\pm1.3 dB$ & $7.8\pm0.0 dB$ 
            & $15.4\pm0.4 dB$ & $9.0\pm0.2 dB$
            & $17.4\pm1.0 dB$ & $13.1\pm1.9 dB$ \\
        512 &$16.5\pm0.3 dB$ & $7.8\pm0.0 dB$
            & $16.8\pm0.4 dB$ & $7.8\pm0.0 dB $
            & $15.6\pm0.6 dB$ & $10.5\pm1.2 dB $
            & $19.5\pm1.7 dB$ & $14.2\pm1.7 dB$ \\
        1024 &$16.6\pm0.3 dB$ & $7.8\pm0.0 dB$
            & $16.4\pm1.0 dB$ & $7.8\pm0.0 dB $
            & $15.6\pm1.1 dB$ & $11.4\pm0.8 dB$ 
            & $19.6\pm1.4 dB$ & $13.7\pm1.7 dB$ \\
        \hline
    \end{tabular}
\end{table*}

\begin{figure}[h!]
    \centering
    \includegraphics[width=\columnwidth]{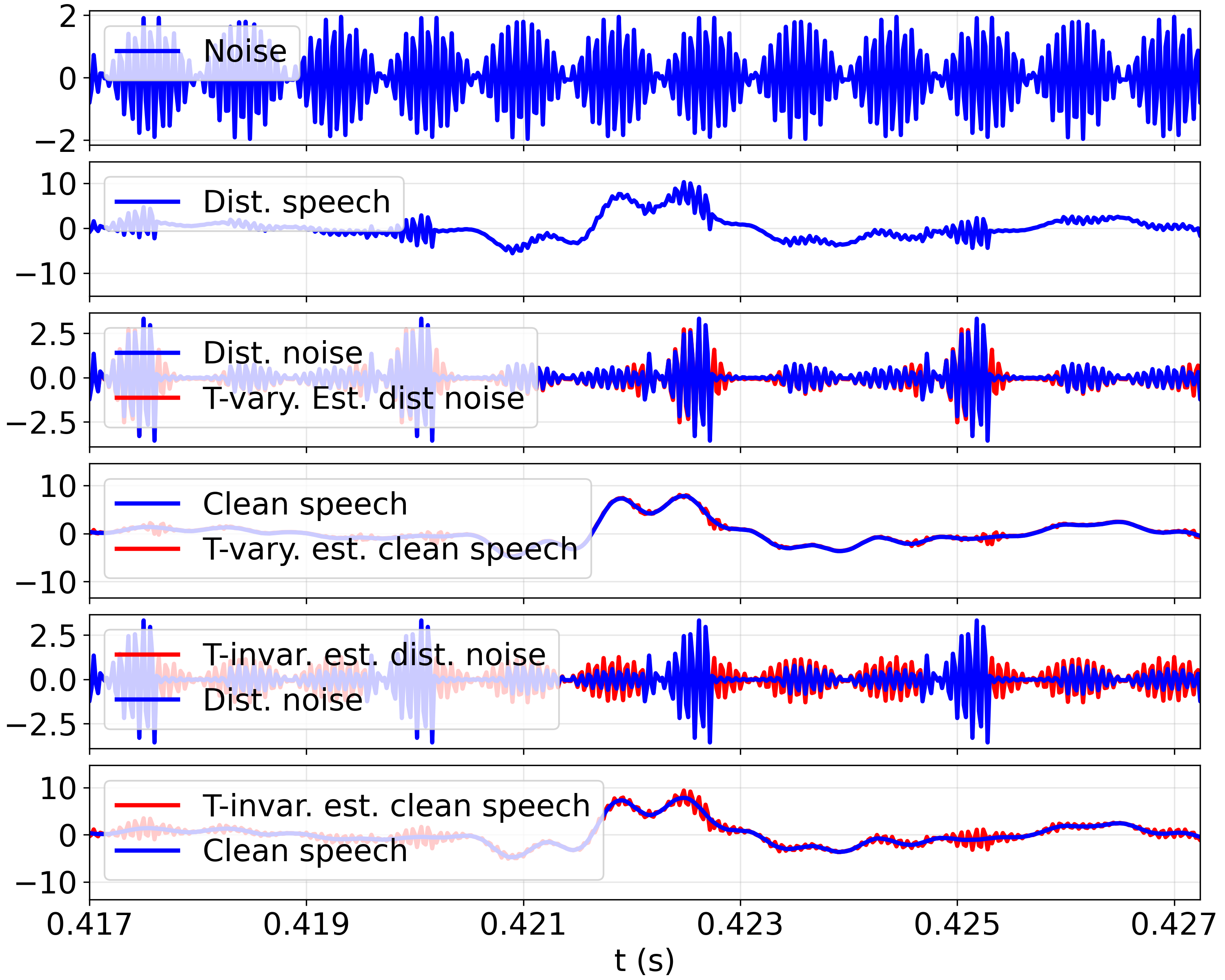}
    \caption{Example sample from the speech-distortion task. The plots show a zoomed-in segment of the $1$s speech sequence.}
    \label{fig:sample_speech_dist_result}
\end{figure}

Table~\ref{tab:vary-Hneurons} compares time-varying and time-invariant models across architectures, with equal parameter counts in each row. The time-varying model consistently outperforms the time-invariant one, reaching up to $19.6$dB vs.\ $14.2$dB SI-SNR from a $5$dB starting point. All time-invariant models plateau at $7.8 \pm 0$dB, consistent with learning an average across modes as visualized in Figure~\ref{fig:sample_speech_dist_result}.

% Observation: Identity; L1 vs L2
With identity activation, adding layers or neurons yields no significant gains for either model.
% Observation: GELU; L1 vs L2
With GELU, two hidden layers consistently match or exceed one layer, suggesting depth is beneficial only with nonlinear activation.
% Observation: L1; Identity vs GELU
Under one hidden layer, GELU improves the time-invariant model across all $\Hneurons$ but offers no consistent gain for the time-varying model.

\subsubsection{Number of parameters:}

%---------------------------------------------------------------------
%---------------------------------------------------------------------

\begin{table}[]
    \centering
    \caption{SI-SNR for speech denoising task for $\Hneurons=512$ and varying $\nbparam$ such that $\Nstate_{invar}=\Nbasefnc\Nstate_{vary}$ and $\Nbasefnc=\Nbasefnc_A=\Nbasefnc_B=\Nbasefnc_C$. }
    \label{tab:vary-Nstate}
    \newcolumntype{n}{>{\centering\arraybackslash}p{3cm}}
    \begin{tabular}{cccc}
        \hline
        $\nbparam$&$\Nstate_{invar}=\Nbasefnc\times\Nstate_{vary}$ & Time-Vary. & Time-Invar. \\
        \hline
        13&$\,4=2\times2$ &$13.6\pm0.6 dB$ & $7.8\pm0.0 dB$\\
        49&$16=4\times4$ &$16.5\pm0.3 dB$ & $7.8\pm0.0 dB$\\
        193&$64=8\times8$ &$19.4\pm0.8 dB$ & $7.8\pm0.0 dB$\\
        \hline
    \end{tabular}
\end{table}

With fixed $\Hneurons$ and increasing parameters per neuron $\nbparam$ (Table~\ref{tab:vary-Nstate}), the time-varying model improves while the time-invariant model is unaffected. This suggests that additional parameters enhance the time-varying model's ability to capture non-stationary dynamics, but cannot compensate for the time-invariant model's lack of inherent flexibility.

%\kern-1em
\subsubsection{Basis functions allocation:}

% Intro
Table~\ref{tab:basis_allocation} examines how to allocate a fixed basis function budget among $\Acoeff, \Bcoeff, \Ccoeff$. 
% Observations
Allocating the entire budget to $\Nbasefnc_A$ performs significantly worse than other configurations, while allocating it fully to $\Nbasefnc_B$ or $\Nbasefnc_C$ yields the best performance, consistent with Section~\ref{sec:results:four_mode_system}. Equal splits across two or three matrices perform similarly to each other.

\begin{table}
    \centering
    \caption{SI-SNR for speech denoising with varying basis function allocation over neuron matrix parameters. $\Nstate_{vary}=4, \Hneurons=512$.}
    \label{tab:basis_allocation}
    \begin{tabular}{ccccc}
        \hline
        Total & $\Nbasefnc_A$ & $\Nbasefnc_B$ & $\Nbasefnc_B$ & Performance \\
        \hline
        12 &4&4&4& 1$6.5\pm0.3 dB$\\
        12 &10&1&1& $\,\,\,8.7\pm0.0 dB$\\
        12 &1&10&1& $17.5\pm0.4 dB$\\
        12 &1&1&10& $18.5\pm0.6 dB$\\
        11 &5&5&1& $16.6\pm0.6 dB$\\
        11 &5&1&5& $16.3\pm0.9 dB$\\
        11 &1&5&5& $15.6\pm1.0 dB$\\
        \hline
    \end{tabular}
\end{table}

\section{Conclusions}
We have proposed time-varying SSM networks parameterized by a dictionary of basis functions with learnable coefficients.
Our investigations illustrate that the proposed models can learn and adapt to periodically switched dynamics, whereas time-invariant SSMs cannot capture such switching behavior.
Extensions to non-stationary settings with non-periodic behavior, as well as exploring different families of basis functions, are important directions for future work. 
Benchmarking against both simpler regression methods and more sophisticated architectures such as non-stationary transformers would further clarify the trade-offs of the proposed approach.

\bibliography{references}   
\appendix

\section{Four mode SLDS}
\label{sec:SLDS_dataset}

\subsection{Model Definition}
\label{sec:four_mode_system_defined_matrices}
We construct an SLDS with four modes, each a linear SSM with $\Nstate=4$, $\nin=1$, $\nout=1$:
\begin{align*}
&\text{System 1: } \Abf_1 = \mathrm{Diag}(0.9, 0.8, 0.9, 0.8), \\
&\hspace{1.8em}\Bbf_1 = \phantom{-}[0.9, 0.8, 0.9, 0.8]^T, \hspace{0.1em} \Cbf_1 = \phantom{-}[0.1, 0.2, 0.1, 0.2].
\end{align*}
\begin{align*}
&\text{System 2: } \Abf_2 = \mathrm{Diag}-(0.1,0.2,0.1,0.2), \\
&\hspace{1.8em}\Bbf_2 = -[0.9, 0.8, 0.9, 0.8]^T, \hspace{0.1em} \Cbf_2 = -[0.5, 0.7, 0.7, 0.5].
\end{align*}
\begin{align*}
&\text{System 3: } \Abf_3 = -\mathrm{Diag}(0.9, 0.8, 0.9, 0.8), \\
&\hspace{1.8em}\Bbf_3 = -[0.1, 0.2, 0.1, 0.2]^T, \hspace{0.1em} \Cbf_3 = -[0.1, 0.2, 0.1, 0.2].
\end{align*}
\begin{align*}
&\text{System 4: } \Abf_4 = \mathrm{Diag}(0.1, 0.2, 0.1, 0.2), \\
&\hspace{1.8em}\Bbf_4 = \phantom{-}[0.1, 0.2, 0.1, 0.2]^T, \hspace{0.1em} \Cbf_4 = \phantom{-}[0.9, 0.8, 0.9, 0.8].
\end{align*}
The system transitions sequentially $(1\rightarrow2\rightarrow3\rightarrow4)$ with equal mode durations.

\subsection{Dataset creation}
\label{sec:four_mode_system:dataset_details}

Each of $(\AbfData, \BbfData, \CbfData)$ can switch or remain fixed independently (Table~\ref{tab:Data_vs_Model_configurations:mse}). 
In the `ooo' case, mode $i$ uses $(\Abf_i,\Bbf_i,\Cbf_i)$. 
In the `xxx' case, all matrices are fixed; there are $4^3=64$ such combinations, over which we report the mean. 
In partial cases such as `oxx', only $\AbfData$ switches while $\BbfData, \CbfData$ are fixed to one of $16$ possible $(\Bbf_j,\Cbf_k)$ pairs. Note that partially switching cases are not subsets of `ooo', so the fully switching system is not necessarily the hardest to predict.

\section{Model and training details}
\label{sec:appendix_info_training}

\subsection{Four mode system}
\label{sec:setup:FourModeSystem}
% Architecture
We use a single hidden layer with $\Hneurons=16$ neurons, one input and one output channel. The time-varying model uses $\Nstate_{vary}=32$, $\Nbasefnc_A=\Nbasefnc_B=\Nbasefnc_C=16$. The time-invariant model uses $\Nstate_{invar}=32$ and $\Nbasefnc=1$ by default. 
% Training
Models are trained with MSE loss for $200$ epochs, batch size $64$, with a linear warmup ($5\%$ of epochs) followed by cosine learning rate decay.

\subsection{Distorted speech}
\label{sec:setup:DistortedSpeech}

% High level model, continue
Architecture choices ($L$, $\Hneurons$, $\Nstate$, $\Nbasefnc$) depend on the task and are specified in the corresponding sections/tables. 
% HPO
The performed Hyperparameter optimization (HPO) uses $200$ random-search trials on a single-layer network with $\Hneurons=512$, identity activation $\Nstate_{vary}=4$, $\Nbasefnc=4$ for time-varying, and $\Nstate_{invar}=16$ for time-invariant, selecting the best configuration by validation performance (Table~\ref{tab:HPO_optim}). 
% Learning
Models are trained with MSE loss for one epoch, with the $48000$-step signal divided into $128$-step segments ($\approx$2.7ms) mimicking an online/adaptive-filtering setting. All metrics are averaged over $10$ runs.

\begin{table}[]
    \centering
    \caption{HPO ranges and chosen values for the distorted-speech task.}
    \label{tab:HPO_optim}
    \newcolumntype{n}{>{\centering\arraybackslash}p{1.2cm}}
    \begin{tabular}{cccc}
    \hline
    Param.&HPO Range&Time-vary. &Time-invar. \\
    \hline
    LR-SSM&$\{10^{-3}, 10^{-4}, 10^{-5}\}$&$10^{-3}$&$10^{-3}$  \\
    LR-others&$\{10^{-2}, 10^{-3}, 10^{-4}\}$&$10^{-2}$&$10^{-2}$  \\
    WD-SSM&$\{0, 10^{-3}, 10^{-4}, 10^{-5}\}$&0&$10^{-5}$  \\
    WD-others&$\{0, 10^{-3}, 10^{-4}, 10^{-5}\}$&$10^{-3}$& 0 \\
    Droprate&$\{0, 0.1, 0.2\}$&0&0  \\
    Batch size&$\{128, 256, 512\}$&128&256  \\
    \hline
    \end{tabular}
\end{table}

\end{document}

%% file: macros.tex
\definecolor{myblue}{rgb}{0.3328, 0.3539, 0.7758}
\definecolor{myblue2}{rgb}{0.0328, 0.0539, 0.4758}
\definecolor{mygreen2}{rgb}{ 0.0328 0.4758 0.0539} 
\definecolor{mygreen3}{rgb}{ 0.0328 0.1758 0.0539} 
\definecolor{myred}{rgb}{0.4758, 0.0328, 0.0539}
\definecolor{myred2}{rgb}{0.75, 0.0328, 0.0539}

\newcommand{\ubf}{\bm{u}}
\newcommand{\xbf}{\bm{x}}
\newcommand{\ybf}{\bm{y}}

\newcommand{\Abf}{\bm{A}}
\newcommand{\Bbf}{\bm{B}}
\newcommand{\Cbf}{\bm{C}}
\newcommand{\CbfBias}{\bm{C}_{bias}}
\newcommand{\Dbf}{\bm{D}}
\newcommand{\R}{\mathbb{R}}
\newcommand{\nout}{n_{out}}
\newcommand{\nin}{n_{in}}
\newcommand{\Hneurons}{h}
\newcommand{\Nstate}{n}
\newcommand{\Wbf}{\bm{W}}
\newcommand{\sbf}{\bm{s}}

\newcommand{\shatbf}{\widehat{\bm{s}}}

\newcommand{\Acoeff}{\bm{A}_{\text{coeff}}}
\newcommand{\Bcoeff}{\bm{B}_{\text{coeff}}}
\newcommand{\Ccoeff}{\bm{C}_{\text{coeff}}}

\newcommand{\Nbasefnc}{K}
\newcommand{\Norm}{\mathcal{N}}
\newcommand{\fo}{\phi}
\newcommand{\nbparam}{p}

\newcommand{\AbfData}{\bm{A}_{D}}
\newcommand{\BbfData}{\bm{B}_{D}}
\newcommand{\CbfData}{\bm{C}_{D}}
\newcommand{\AbfModel}{\bm{A}_{M}}
\newcommand{\BbfModel}{\bm{B}_{M}}
\newcommand{\CbfModel}{\bm{C}_{M}}